\documentclass{cys}

\usepackage[utf8]{inputenc}

\usepackage[russian,english]{babel}  

\usepackage{graphicx}

\usepackage{epstopdf} 
\usepackage{epsf}
\usepackage{adjustbox}
\usepackage{hyperref}
\usepackage{amsmath}
\usepackage{caption}
\usepackage{color}
\usepackage{tcolorbox}
\usepackage{multirow}

\addto\captionsenglish{%
}

\addto\captionsspanish{%
}

\title{Modeling Coherency in Generated Emails by Leveraging Deep Neural Learners}

\author{Avisha Das and Rakesh M. Verma}

\affil{ 
University of Houston \\ Department of Computer Science \authorcr   
Houston, TX 77204, USA            
\authorcr \authorcr
adas5@uh.edu, rverma@uh.edu
\authorcr  \authorcr
}

\begin{document}

\maketitle

\renewcommand{\tablename}{Table}

\begin{abstract}
Advanced machine learning and natural language techniques enable attackers to launch sophisticated and targeted social engineering based attacks. To counter the active attacker issue, researchers have since resorted to proactive methods of detection. Email masquerading using targeted emails to fool the victim is an advanced attack method. However automatic text generation requires controlling the context and coherency of the generated content, which has been identified as an increasingly difficult problem. 
The method used leverages
a hierarchical deep neural model which uses a learned representation of the sentences in the input document to generate structured written emails. We demonstrate the generation of short and targeted text messages using the deep  model. The global coherency of the synthesized text is evaluated using a qualitative study as well as multiple quantitative measures. 
\end{abstract}

\begin{keywords} 
Bidirectional LSTM, Doc2Vec embeddings,
Proactive defense, natural language generation (NLG).
\end{keywords} 






\section{Introduction}
\label{sec:introduction}
Adversarial learning is a major threat to the field of computer security research. With the advancement in technology, the growing dependency on the Internet has exposed users to serious cyber threats like phishing and pharming. Despite considerable research to counter such threats, staggering numbers of individuals and organizations fall prey to targeted social engineering attacks incurring huge financial losses. 

Although attackers change their strategies, previous research~\cite{caputo2014going} has shown that electronic mails (emails) are a popular attack vector. Emails can be embedded with a variety of malign elements~\cite{drake2004anatomy} like poisoned URLs to malicious websites, malware attachments as well as executables, documents, image files, etc. 
Anti-Phishing Working Group (APWG) reports over 270,500\footnote{http://docs.apwg.org/reports/apwg\_trends\_report\_q3\_2018.pdf} unique phishing email campaigns received in the $3^{rd}$ quarter of 2018, rising from a total of around 233,600\footnote{http://docs.apwg.org/reports/apwg\_trends\_report\_q4\_2017.pdf} unique reports identified in the $4^{th}$ quarter of 2017. Phishing reports also reveal the consistent rise in phishing attacks targeted towards financial institutions like payment processing firms and the banking sector. The statistics demonstrate how the threat is worsening as attackers continue to devise more sophisticated (and maybe more effective) ways of scamming victims.

Innovative and unseen attack vectors can trick pre-trained classification techniques~\cite{sommer2010outside}, thus placing the victim at risk. In \textit{email masquerading} attacks, attackers after compromising the email account of an individual can carefully construct a fraudulent email then sent to the contacts known to the compromised individual. This has serious implications, because the attacker has gained uninterrupted access to the inbox, outbox and other private details of the compromised person. Thus by exercising caution, the attacker can emulate the content and context of the emails written by the individual and can communicate with his contacts as a legitimate entity, successfully evading detection and causing harm to the victim.

However, construction of the perfect deceptive email requires fine-tuning and manual supervision. While a fake mail constructed manually by an attacker can guarantee a higher chance of success, the process is both time and labor intensive. In contrast, an automated text generator can be trained to synthesize targeted emails much faster and in bulk, thereby increasing the odds of a successful attack. However, the bottleneck in this case, lies in whether the system can generate high quality text, free from common flags like misspellings, incorrect and abusive language, over-usage of action verbs, etc., which can be picked up by a classifier easily. Thus, proactive research in this area of deception based attacks using email masquerading techniques requires further sophisticated experimentation.   

Advances in the field of natural language processing have introduced newer and  sophisticated algorithms which enable a machine to learn and generate high-quality textual content on a given context. Grammar based tools like the Dada Engine~\cite{baki2017scaling}, N-gram language models~\cite{chen2014two} as well as deep neural learners \cite{yao2017automated} have been used to study and replicate natural language based attacks. The aim is to facilitate proactive research by predicting newer attacks and reinforce against such unseen yet impending threats. 

At the hands of an attacker, language generation techniques can become dangerous tools for deception. With access to proper training data, deep learning neural networks are capable of generating textual content. This property has been leveraged by researchers for generating tweets~\cite{sidhaye2015indicative} and poetry~\cite{ghazvininejad2016generating},~\cite{xiedeep}, etc. While limited, proactive research has been pursued by using deep learners for generation of fake reviews~\cite{yao2017automated}, grammar based techniques~\cite{baki2017scaling} as well as simplistic deep networks~\cite{DAS18.4} have been leveraged for email generation. Thus, we can assume that it is not long before phishers and even spammers resort to such techniques to generate newer kinds of malicious attack vectors. 

Following a proactive mode of study, we identify the underlying implications of how an automated machine learning technique, here, deep learners can be leveraged to synthesize email bodies for the purpose of email masquerading attacks. 
Along with demonstrating the systems' performance using qualitative and quantitative methods, we study the effectiveness and practicality of such systems by comparing a hierarchical deep network with a baseline word prediction model. Our key contributions are as follows:

\begin{itemize}
\item A collection of the most common signals that set apart a malicious email from its legitimate counterpart (Section~\ref{sec:textcues}) as mentioned in previous literature~\cite{verma2012detecting},~\cite{drake2004anatomy},~\cite{ferreira2017phish} on analyzing phishing emails. We focus on cues that are more prevalent in email bodies for our evaluation. This is necessary to observe the quality of the system generated emails.
\item Leveraging deep neural networks for generating targeted email bodies 
(Section~\ref{sec:genarch}). While generation of coherent emails is challenging~\cite{DAS18.4}, we use a hierarchical network that consists of two stages - an architecture which uses a word prediction model to generate probable candidate sentences which are then passed onto a sentence selection model, based on distributed vector representations of the email content, to select the best possible set of sentences. Such a two-staged architecture should be suitable for generating longer content while maintaining coherency.

\item Comparing the performance of the hierarchical system with a baseline word prediction based model by using multiple quantitative and qualitative metrics (Section~\ref{sec:syseval}). Additionally, we analyze the effectiveness of our system by measuring the \textit{syntactical correctness}, \textit{coherency}, \textit{fluency} and \textit{legitimacy} of the fake emails by conducting a human evaluation. 
\end{itemize}

\section{Background}~\label{sec:prelim}
This section presents a list of cues usually observed in spoofing emails that demarcate such attack vectors from their legitimate counterparts. Highlighting and studying such common signals helped us prepare, process and evaluate our training data as well as the generated emails. 
The goal of the method is defined after studying the common features in malicious emails, followed by a detailed description and demonstration of the baseline word prediction and the hierarchical sentence selection models used for the generation task.


\subsection{Textual features in spoofing emails}~\label{sec:textcues}
Use of textual features, like presence of common action words, organization names, poisoned links to malicious webpages of financial institutions, grammatical errors, etc., is common in phishing email detection methods~\cite{verma2012detecting},~\cite{verma2012two},~\cite{verma2013semantic}. Moreover, researchers have widely studied spam, phishing, spear phishing emails to identify common signs that appear across malicious emails~\cite{drake2004anatomy},~\cite{ferreira2017phish}. However, since such signals are certain signs of malign intent an attacker would consciously avoid incorporating these words in a targeted email. Assuming this in mind, the generator should also learn to identify and eliminate overuse of such words. 

Therefore, we curate a list of textual cues frequently used in spoofing emails after careful review of phishing email literature~\cite{drake2004anatomy},~\cite{verma2012detecting},~\cite{verma2012two},~\cite{verma2013semantic},~\cite{ferreira2017phish},~\cite{chandrasekaran2006phishing}. The list of these textual cues along with examples have been provided in Table~\ref{tab:phishcues}. Researchers prefer to train their proposed detection methods on publicly available data. Since phishing emails are fairly rare, we base our evaluation on the largest publicly available dataset of malicious emails:   Nazario Phishing 
Corpus.\footnote{ https://monkey.org/\textasciitilde jose/phishing/} The Base-64 encoded HTML content in the emails are filtered out and finally 3,392 fairly clean emails with textual content ($>$10 words) are used to extract the enlisted spoofing cues. 


\begin{table}[!htb]
\caption{Common Spoofing Cues in Phishing Email bodies}
\label{tab:phishcues}
\centering
 \resizebox{1\columnwidth}{!}{ 
\begin{tabular}{|l|l|}
\hline
\textbf{Feature Types                                                               }& \textbf{Examples}                                                                                                                                                                                                                                                                                                                                  \\ \hline
\textbf{Organization Names}                                                          & \begin{tabular}[c]{@{}l@{}}(a)~\textit{Financial} like eBay, PayPal, Bank of America,\\ Western Union\\(b)~\textit{Government} like Internal Revenue services, \\United Parcel Service\\(c) \textit{Software} like Dell, Microsoft, Apple\end{tabular}                                                                                                                               \\ \hline
\begin{tabular}[c]{@{}l@{}}\textbf{Action Verbs and} \\ \textbf{Urgency Adverbs}\end{tabular} & \begin{tabular}[c]{@{}l@{}}(a)~\textit{Action verbs} like click, follow, visit, go, update,\\ apply, submit, confirm, cancel, dispute, enroll, \\login, answer, reply\\(b) \textit{Adverbs implying urgency} like today, instantly,\\ straightaway, straight, directly, once,\\ urgently, desperately, immediately, soon, shortly,\\ presently, before, ahead, front\end{tabular} \\ \hline

\begin{tabular}[c]{@{}l@{}}\textbf{Persuasion Principles}\end{tabular} & \begin{tabular}[c]{@{}l@{}}(a)~\textit{Authority} like an email from a bank asking the\\ victim to update password of his online account\\ (b)~\textit{Social proof} denoted by Emails from \\the IT department of the target's institution\\(c) \textit{Distraction} using emails where a target\\ is tempted to click a link in order to\\ receive a prize \\(d)~\textit{Reciprocation} appealing the victim to respond\\ like resetting a password or paying a bill\\ by clicking a link to a fraudulent website \end{tabular} \\ \hline

\textbf{Misspelled Words}                                                            & \textit{Typographical errors} like Paypl, Bnk Amrica, etc.                \\ \hline
\textbf{Presence of Links  }                                                         & \begin{tabular}[c]{@{}l@{}}URLs to malicious websites \\ like https://www.maybank2u.com.my, etc. \end{tabular}                                                                                                                                                                                                                                                                         \\ \hline
\textbf{Other languages}                                                      &\begin{tabular}[c]{@{}l@{}} \textit{Non-English words} like Aviso Importante de BBVA,\\ societe, Transaktionen   \end{tabular}                                                                                                                                                                                                                                                                  \\ \hline
\end{tabular}}
\end{table}

While machine learning systems can detect common cues, these detectors largely depend on historical data. To keep up with advanced reinforcement techniques, a phisher also resorts to employing sophisticated techniques for making their attacks more targeted to increase rate of success. Thus, for social engineering based attacks like email masquerading, spear phishing, or targeted phishing, an attacker may choose to avoid such easily identifiable red flags while generating fake emails. Therefore, in our proactive study, we also refrain from overuse of such spoofing cues in the synthesized emails. 

\subsection{Task Description}
Email masquerading is steadily growing into a serious cybersecurity threat and has started gaining much attention from security researchers. This paper aims at providing a proactive paradigm to this issue. Some of the key questions are: Given a large dataset of manually written emails, can an automated system learn to emulate the writing style of a human? Is the generated email content coherent and syntactic? Can an individual differentiate between a system generated email and a manually written one?

The hierarchical model is compared with a baseline word generation architecture. Furthermore, the system performance is studied using multiple quantitative metrics along with a qualitative evaluation conducted through a human survey.

\subsection{Architecture for Text Generation} \label{sec:genarch}

Textual content can be considered as a sequence of words and characters placed together to convey meaningful information. In the realm of text generation, deep neural architectures have seen unprecedented success in emulating one's writing when trained on huge amounts of written textual content~\cite{yao2017automated},~\cite{ghazvininejad2016generating},~\cite{sutskever2011generating}. 

Recurrent Neural Networks (RNNs) are capable of retaining information learned from text sequences -- helpful for learning representations of word sequences in the input text. The trained language model can subsequently generate samples similar in form and context to the input data. We leverage this ability of RNNs for our proactive protection scheme - generation of targeted emails suitable for spear-phishing or masquerading attacks. 
Moreover, Long Short Term Memory (LSTM) Networks, an improved version of RNN, have proved to be better at handling dependencies in longer sequences of text~\cite{li2016deep},~\cite{yao2017automated}. 

The architectures use \textit{words} as units for generation~\cite{xiedeep},~\cite{henderson2014word} by leveraging LSTMs as building blocks for learning the language model. We use a hierarchical model that merges sentence selection with iterative text generation to generate the best set of human readable samples. We compare the architecture, sampling and generation phases for these two architectures followed by an evaluation of their performance.

\subsubsection{Training a Word Prediction Model.}\label{sec:wordpred}
\begin{figure*}
    \centering
    \begin{minipage}{.5\textwidth}
        \centering
        \includegraphics[width=.7\linewidth,height=.7\linewidth]{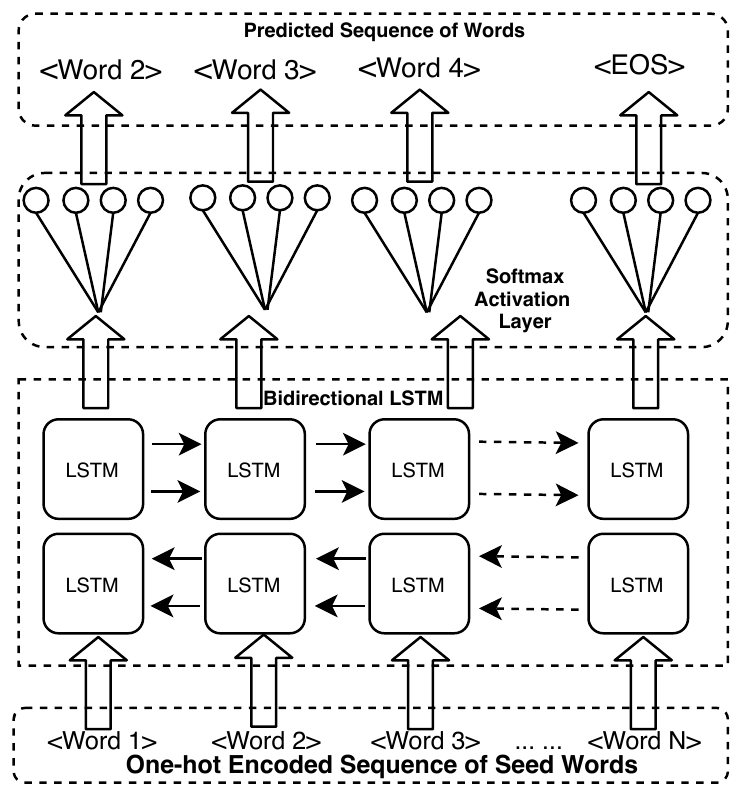}
        \caption{Word generation model with Bi-LSTMs}
        \label{fig:model_1}
    \end{minipage}%
    ~
    \begin{minipage}{.5\textwidth}
        \centering
        \includegraphics[width=.7\linewidth,height=.7\linewidth]{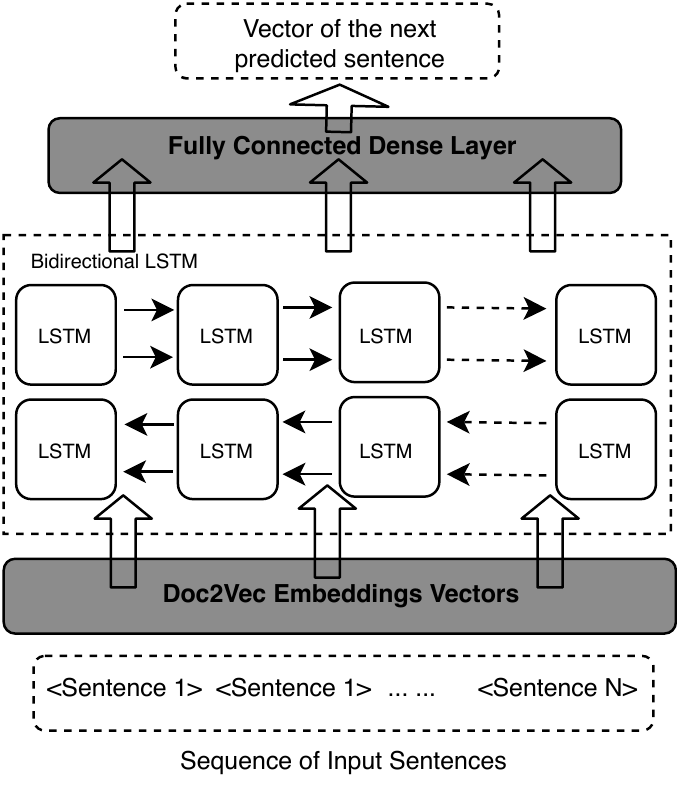}
        \caption{Sentence Selection model using Doc2Vec with Bi-LSTMs}
        \label{fig:model_2}
    \end{minipage}
\end{figure*}

Our first model is a straightforward word-based language model built using RNNs~\cite{xiedeep}. In our implementation, we use Bidirectional LSTMs (Bi-LSTMs)~\cite{Graves2005FramewisePC} as the network to build the model. 
Figure~\ref{fig:model_1} shows the overall model for word prediction. We describe the training and generation phases for this baseline generation architecture.

\textbf{\textit{Training Phase.}} 
We use Bidirectional LSTMs (Bi-LSTMs) as our network for text generation model. The input to our text generation system are one-hot vector sequences of words ($w_{0}$, $w_{1}$, $w_{2}$, ..., $w_{N-1}$) where $N$ is the sequence length. At every time step $t$ (starting with 0), we feed a word ($w_{t}$) into the hidden layer which predicts the next word $w_{t+1}$. 
Through error optimization, the model learns to capture the best representation of coherent word sequences that constitute the textual content to be generated - in this case, the body of an email.


\textbf{\textit{Generation Phase and Temperature Regulation.}} During the generation phase, we feed a sequence ($N$) of seed words ($W_{seed_0}, W_{seed_1}, W_{seed_2}, ..., W_{seed_N}$) into the trained Bi-LSTM model, used to start off the word generation system. When the model gets a seed word ($W_{seed_0}$) as input, it outputs the next word ($W_{1}$) by selecting the one most likely to occur after $W_{seed_0}$ depending on the conditional probability distribution, $P(W_{1}|W_{seed_0})$. 
When this aforementioned step is extended for an input sequence of $N$ seeds, the model (Figure~\ref{fig:model_1}) can generate a text body of $N+1$ words, the $N+1^{th}$ word being the output.

The final layer of the model, which calculates the above conditional probability, is a $softmax$ normalization which is used for computing the distribution for the next word followed by subsequent sampling. 
We use $temperature (\tau)$ as the hyper-parameter for selecting our word samples - regulating the parameter $\tau$ in Equation~\ref{eq:softmax} encourages or controls the diversity of the generated text. The novelty or eccentricity of the generative model can be evaluated by varying the $temperature$ parameter  between $0 < Temp. \leq 1.0$. While, lower values of $\tau$ generate relatively deterministic samples, higher values can make the process more stochastic. 

Equation~\ref{eq:softmax} shows the probability distribution built by the model for the sequences of words along with the incorporation of \textit{temperature} control parameter($\tau$)
$P(W_{t+1}|W_{t'\leq t}) = softmax({W_{t}})$, 
\begin{equation}\label{eq:softmax}
P(softmax(W_{t}^{j})) = \frac{e^{\frac{W_{t}^{j}}{\tau}}}{\sum_{j=1}^{n}e^{\frac{W_{t}^{j}}{\tau}}}     
\end{equation}

\subsubsection{Hierarchical Sentence Prediction Model.}
Controlling global coherency and structure in automated text generation is a non-trivial task. Using a straightforward word prediction model makes it increasingly difficult to control the quality as well as coherency of the generated text. We use an hierarchical model~\cite{D2Vlink} consisting of two stages - generation of sentence candidates\footnote{sequences of words highly likely to appear together in a sentence in email body} followed by sentence selection model. For the sentence selection model, we use Doc2Vec~\cite{le2014distributed} embeddings to learn representations of the email bodies, where each email is treated as a document. Given a set of sentences as starting point, we describe how the trained Doc2Vec model is used to generate a new sentence.

\textbf{\textit{Sentence Selection using Doc2Vec.}} Use of embeddings has been regarded as a favorable way to represent context in a piece of text, either in the form of word phrases, sentences or paragraphs. Doc2Vec~\cite{le2014distributed} can effectively learn the numeric representation of a paragraph or even a document, irrespective of its length. 
In this model, we use Doc2Vec to learn better representations of the sentences in a piece of text - for example, the body of an email. Figure~\ref{fig:model_2} shows the stages of the sentence selection model. 
The model starts with training a Doc2Vec model on the entire document. The goal is to predict the next sentence (here, its vector representation) given a sequence of sentences as starting point.  As shown in Figure~\ref{fig:model_2}, using the trained Doc2Vec model, we vectorize a set of seed sentences and then feed the vectors to the Bi-LSTM layer in order to learn the model for sentence prediction. 

\textbf{\textit{Generation of Sentence Candidates.}} The first stage deals with the generation of a set of candidate sentences and/or phrases. The input to the whole model is a set of seed sentence sequences. The first level of the network is a trained word prediction model which takes as input the last $N$ words from the given seed sentences. The model architecture is same as the one described in Section~\ref{sec:wordpred}. This sequence of $N$ words is used to generate a set of candidate sentences which are transformed into their Doc2Vec vector representations for comparison and selection.

\textbf{\textit{Generation of newer sentences.}} The seed sentences are fed into the trained Bi-LSTM-based Doc2Vec model for sentence selection. This constitutes the second stage of the hierarchical architecture and selects the best sentence vector depending on the input sequences. The generated vector is then compared with the selected candidate sentences - output of first level using a cosine similarity function. The most similar candidate is then produced as output of the hierarchical generation model.  

\textbf{\textit{Generation phase and Temperature Regulation.}} The parameters used and the purpose for temperature regulation during the generation phase using the hierarchical model is similar to the generation phase explained in Section~\ref{sec:wordpred}. The selected best sentence from the generated candidate sentences or phrases are merged with the sequence of input seeds and fed into the model to generate newer sentences.

\begin{figure*}[!htb]
\centering
\includegraphics[width=\linewidth,height=.5\linewidth]{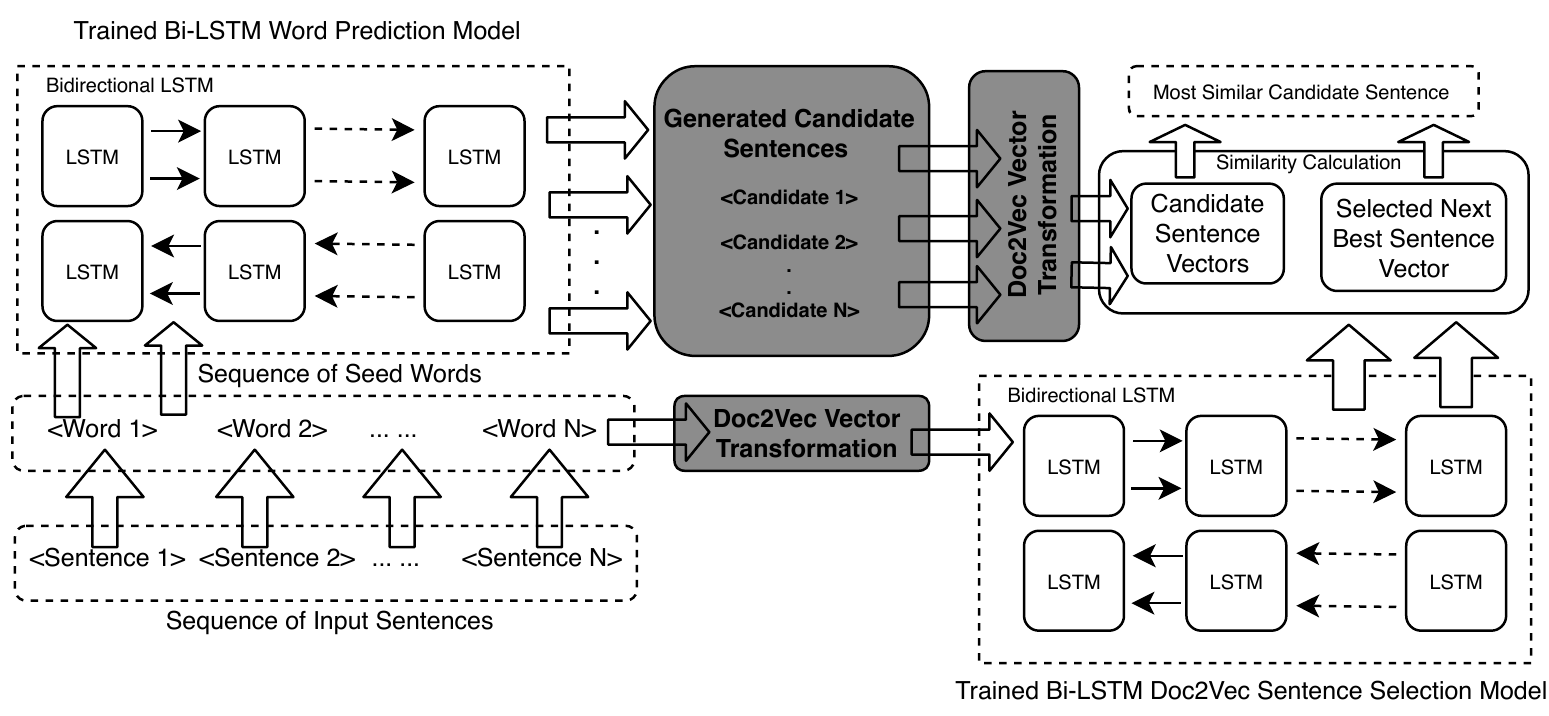}
\caption{Word prediction with sentence selection using Bidirectional LSTMs}
\label{fig:rnnmodel_3}
\end{figure*}

Therefore, the complete \textit{hierarchical} or multi-stage architecture has been shown in Figure~\ref{fig:rnnmodel_3}. The sequence of sentences, called \textit{seed sentences} ($Sent_{1}$, $Sent_{2}$, ..., $Sent_{N}$), are first chosen as input to be given to two pre-trained models - one, a word-based language model and the other a model trained on Doc2Vec embeddings of sentences in documents. 
The word-based language model uses a sequence of the last $N$ words from the given sentences as input. It then outputs the most probable word to appear in the $N+1^{th}$ position based on the trained model. The model repeats this step in a feedback setup to generate sequences of sentences. These generated sentence sequences are called \textit{candidate sentences}. We refer to these generated candidate sentences as $Sent_{cand1}$, $Sent_{cand2}$, ..., $Sent_{candX}$.\footnote{$X$ refers to the number of candidates to be generated.} 


The seed sentences ($Sent_{1}$, $Sent_{2}$, ..., $Sent_{N}$) are converted to their Doc2Vec embedding vectors -- $Sent_{d2v1}$, $Sent_{d2v2}$, ..., $Sent_{d2vN}$ -- using the trained Doc2Vec vector transformation model. These sentence vector representations are then fed into the BiLSTM-based sentence selection model which had been trained on sentence embeddings using Doc2Vec representations (Figure~\ref{fig:model_2}). Given a sequence of sentence embeddings, this model selects the most suitable sentence vector to follow the given sequence. 

In the final step, we compare each generated candidate sentence to the selected sentence vector. To explain this step, we select the candidate sentence $Sent_{cand2}$ and convert it into its Doc2Vec representation ($Sent_{cand-d2v2}$). The selected sentence vector from the Doc2Vec sentence selection model given the $N$ seed sentences is $Sent_{d2vN+1}$. We use the Cosine similarity metric to calculate the similarity between the two vectors $Sent_{cand-d2v2}$ and $Sent_{d2vN+1}$. This step is repeated for all the candidate sentences and the sentence candidates with the highest similarity values are then chosen as the output of the model.

\section{Data Collection and Setup} ~\label{sec:setup}
A large amount of high quality data is required for text generation. This becomes crucial when automating the composition skill and pattern of an individual. Here, we describe the source and collection steps of the data used in the email body generation task.

\subsection{Data Collection}

To produce the best sample to emulate a targeted attack, the trained generation model must synthesize an email body similar to an email composed by the individual whose writing style it is trying to emulate. Thus, for training our model we make use of the largest publicly available source of legitimate emails, the Enron Corpus~\cite{enron}. Table~\ref{tab:ldata} summarizes the statistics about our dataset. 
We evaluate the average number of sentences, average vocabulary as well as average number of words in a typical email body.
The step-wise data collection process along with the data preparation and pre-processing steps are given below.


\textbf{Building the dataset of legitimate emails.} We use the largest publicly available dataset of benign emails - Enron  Corpus.\footnote{\url{https://www.cs.cmu.edu/~enron/enron\_mail\_20150507.tar.gz}} Since we aim at synthesizing the writing style of humans, it is necessary to make the process of email masquerading more effective. For building our training and evaluation dataset, we make use of `legitimate' emails which we collect based on the following assumptions.  We assume that the emails that an individual, within Enron, receives are legitimate i.e., they have been identified as benign by the mail server. Also, the emails that the individual sends out are also benign. The Enron corpus consists of a large number of emails ranging from spam, deleted advertisements, etc., which is useless for training purposes. Hence, for our purpose, we use emails from two types of email directories - the Inbox (Received folder) and Outbox (Sent folder) of the individuals within Enron corpus. 

The collection process involved the following steps: (a) We use the collection of 517,401 emails from the publicly available Enron corpus; (b) We choose the emails that are composed of a minimum of 10 words and contain a minimum of one sentence; (c) We also calculate the average statistics of the dataset to determine the generic length, number of words, vocabulary size of the emails we plan to synthesize.  

\begin{table}[!htb]
\caption{Legitimate Data Statistics}
\label{tab:ldata}
\centering
\begin{tabular}{|l|r|}
\hline
\textbf{Attributes}        & \textbf{Values}                     \\ \hline
Dataset Size                         & 517,401                          \\ \hline
Total Number of Words                & 167,692,695                      \\ \hline
Avg. Number of Words                 & 324                              \\ \hline
Total Vocabulary                     & 2,028,671                        \\ \hline
Avg. Vocabulary                      & 143                              \\ \hline
Avg. Sentence Length                 & 25                               \\ \hline
Total Number of Sentences            & 6,590,091                        \\ \hline
Avg. Number of Sentences             & 13                               \\ \hline
\end{tabular}
\end{table}

\subsection{Assumptions and Data Preprocessing}
We assume a typical scenario where the attacker has compromised and gained access to the mail account, all email communication, personal details, email contact list, etc.

Our primary focus is on the automated generation of textual content of the body of an email. Hence, we do not account for the generation or evaluation of header information in an email. Moreover, we also do not consider the generation of email attachments, links or email addresses that maybe present in the body. The goal is to make the synthesized email content more generic to allow inclusion of personalized information like named entities, location names, etc., depending on the communication between the victim and the compromised individual. Also, from the architectural viewpoint, including named entities, personalized details, etc., is unnecessary during training phase and will eventually blow up the vocabulary of our word-based architecture, therefore confusing it. 
To avoid the inclusion of unnecessary word tokens and to generalize the context, we replace the named entities in the email bodies with the $ent$ tags - this includes replacing person names as well as locations (if applicable). We use the Entity Recognizer Module implemented in SpaCy for Python 3.6 for entity tagging and replacement.

In most cases, the body of the Emails contain elements which are of little or no value to our training and evaluation process. We apply the following pre-processing steps to clean our data while making sure not to delete any important information:
\begin{itemize}
\item We remove all trailing spaces, newline characters, etc., from the text body
\item Replacement of named entities (person, location, etc.) with $ent$ tag
\item We replace Email addresses in the body with $emailID$ tag 
\item Replacement of URL links with the $link$ tag
\item Adding the \textit{Start of Text} ($<$SOT$>$) and \textit{End of Text} ($<$EOT$>$) tags to the email bodies to respectively mark the beginning and end of the content.
\item Sanitization of non-ASCII characters from the text 
\item Removal of HTML text fragments and broken links from text body
\item We also lowercase our textual content and remove special characters like \#, \$, \%, \@, etc.
\end{itemize}

\subsection{System Setup and Methodology} 
The system was developed in Python 3.6 using Keras (Version 2.2.4) and TensorFlow (Version 1.11.0). In our experimental evaluation, the LSTM network consists of 128 hidden units, since we are using Bi-LSTMs, the total number of states is 256.  
 We consider an unrolling size of 15, i.e., the network looks back up to a sequence of 15 words to predict the next probable word. 
Among the other hyperparameters, we consider a batch size of 50 with a learning rate $10^{-2}$. Each architecture was trained for a total of 50 epochs for building the language model. 

The above  set of hyperparameters has been chosen based on our empirical evaluation as explained in this section. We considered a hyperparameter optimization scheme where we look at the GridSearch technique provided by the Scikit-Learn library. We tune the networks for batch sizes 35, 50, 75 and epochs of 10, 20 and 50. We also experiment with LSTM hidden units of 128 and 256. We choose these values since the training time is much more reasonable. We manually selected the unrolling size length by experimenting with values between 10 to 25 in increments of 5 units. Similarly, the learning rate ($10^{-2}$) was selected after running the algorithm with learning rates of $10^{-3}$, $10^{-2}$ and $10^{-1}$.  

All our experiments were conducted on a server with 4 Tesla M10 GPUs using CUDA (Version 9.1.85) with a 3.20GHz Xeon CPU E5-2667 and 512 GB of memory.  
We use \textit{Adam}~\cite{kingma2014adam} algorithm for the gradient optimization in the word-based as well as the Doc2Vec language modeling architectures. For the calculation of the loss functions, we use \textit{categorical crossentropy} for the word-based model and a regression based loss function, \textit{Log-Cosh} for learning the model for Doc2Vec embeddings.  

For the word prediction model, the starting seed of 15 words were chosen randomly from a set of starting sentences in the email bodies. For generating text using the hierarchical architecture, the system was given a set of candidate sentences comprising of the first sentence of 10 randomly selected legitimate emails from the Enron corpus. 

\section{System Evaluation and Analysis} \label{sec:syseval}
This section describes the experimental setup for the training and generation phases of the baseline and deep generation model. The evaluation setup for analyzing system performance based on the quantitative and qualitative nature of the generated content as well as their practical implications have been discussed in detail.


\subsection{Evaluation Setup} For our evaluation setup, we use a total dataset of 517,140 emails. We set apart 5\% of the dataset for validation during training and the rest was used for training the model. The separate training and validation subsets are crucial for determining the performance of our prediction model on unrelated data. We train the deep generative models each for 50 epochs with the validation of the model performance after every epoch, finally selecting and saving the model that achieves the best performance on validation data. 

\subsection{Evaluation Metrics}
We evaluate the performance of the generative model along three dimensions: (a) a \textit{quantitative} evaluation using a word-gram based measure of overlap and uniqueness between the synthesized and the targeted email bodies as well as model perplexity; (b) a simple \textit{qualitative} evaluation where we inspect the samples generated by the generative word-based and hierarchical models by selecting some samples; and finally, (c) a \textit{human} evaluation which consists of a survey with 6 participants providing feedback on the overall syntax, coherency, and fluency of the synthesized email bodies as compared to their legitimate counterparts. 

\subsubsection{Quantitative Evaluation}
Measure of model \textit{perplexity} is an established method to determine the ability of a language model. We compare the perplexity of the word-based model (Generator$_{word}$) with the deep hierarchical sentence selection model (Generator$_{sentence}$). A lower perplexity ensures a more stable predictive ability. Model perplexity is defined as:
\begin{equation}
    PPL = 2^{\frac{NLL}{T}}
\end{equation}
Here, $NLL$ refers to negative log likelihood of the model and $T$ is the length of the training sequence for which the model perplexity is measured. A more detailed definition for a word-based generation model:
\begin{equation}
  PPL = 2^{-\frac{1}{T}\sum_{t=0}^{T-1}log(P(W_{t+1}|W_{0}...W_{t}))}  
\end{equation}
During our second phase of quantitative evaluation, we use a novel metric to observe the semantic coherence across the generated samples. Coherency in an email should account for the adequacy in information between the sentences or even words in the email body~\cite{li2016deep}. Taking into account the mutual information between a word and its predecessors in the sentence to calculate semantic coherence, we can define the following measure:
\begin{equation}
\begin{split}
    Coherence = \frac{1}{T} \sum_{t} (\log p_{fwd}(w_{t}|w_{t-1}) +\\
                            \log p_{bwd}(w_{t-1}|w_{t}))
\end{split}
\end{equation}
The mutual information between two words acts as a measure of how likely it is for the words to appear together in the context, thus a higher value of mutual information means better coherence~\cite{guo2003using}. We take into account both the forward as well as the backward probabilities of occurrence of the bigrams ($w_{t}, w_{t-1}$). To control the influence of the length of the generated text, the values are scaled by the total length (i.e., number of words) of the generated sentence/sequence.

Measuring perplexity can provide a false sense of language model's performance - while a low value denotes the model's ability to replicate the input text; it may emulate the characteristics of the input text too much. To measure the overlap between the generated and targeted email text, we consider the ratio of common word-N-grams between the ground-truth (or legitimate) email and its `fake' counterpart. For evaluation purposes, we select $N=3$, i.e., trigram overlap.

\begin{table}[!htb]
\caption{Quantitative evaluation using generated emails}
\label{tab:pplval}
\centering
\resizebox{\columnwidth}{!}{ 
\begin{tabular}{|c|r|r|r|}
\hline
\textbf{Model} & \textbf{Perplexity} & \textbf{Coherence} &\begin{tabular}[c]{@{}c@{}}\textbf{\textbf{Trigram}} \\ \textbf{\textbf{Overlap} (\%)}\end{tabular}\\ \hline
Generator$_{word}$ &  8.97 & 2.8 & 43.7\\ \hline
Generator$_{sentence}$ & 4.59 & 4.9 & 66.8\\ \hline
\end{tabular}}
\end{table}

The results of the evaluation have been presented using a dataset of 100 generated emails. The coherence measure has been calculated based on the language model built using the Brown Corpus\footnote{https://www.nltk.org/book/ch02.html} available with NLTK package in Python. We use a list of top 1000 most common trigrams in the legitimate emails dataset for the calculation of our trigram overlap. We observe an increased measure of coherence as well as occurence of common trigrams in the emails generated by the sentence based models.


\subsubsection{Qualitative Evaluation}
The most simplistic form of qualitative evaluation is observing the nature of randomly selected samples from the generated text. We have defined common malicious cues in Section~\ref{sec:textcues}. Drawing motivation from Table~\ref{tab:phishcues}, we inspect the generated email content for signs of obvious malign intent as well as other prominent features which can be used to evaluate our system performance.

Here, we include two samples each, generated using the word-based model and the hierarchical model at different temperatures for the purpose of our analysis. One noticeable feature of the emails generated by the systems is the number of sentences in the samples. While the more genuine looking samples selected from the outputs of the Generator$_{sentence}$ model tend to be longer i.e., consists of more than one sentence, the samples from the baseline model (Generator$_{word}$) are comparably more terse.
Also, longer sequences from the deep generation model generated at a higher $\tau$ tend to include more $ent$ tags at random places within the text.

\textbf{(A) Samples generated by Generator$_{word}$:}
\begin{tcolorbox}
\textbf{A1. Sampling at $\tau$ = 0.9:} $ent$ Could you give me a moment for why I would appreciate everyone's support . $ent$ Best wishes, $ent$ \\
\textbf{A2. Sampling at $\tau$ = 0.5:} Hi , I hope you are having a good $ent$ interview. $ent$
\end{tcolorbox}

\textbf{(B) Samples generated using Generator$_{sentence}$:}
\begin{tcolorbox}
\textbf{B1. Sampling at $\tau$ = 0.5:} Yes . We had $ent$ them to set up a date . $ent$ We are now in London on Monday for Christmas.\\
\textbf{B2. Sampling at $\tau$ = 1.0:} Hi ? $ent$ What do you have the steps or makes it a minute. $ent$ P.S. You can $ent$ email $ent$ with changes to a reports that $ent$ has delivered several to the members with the office
\end{tcolorbox}

\textbf{\textit{Human Evaluation.}} While automated qualitative and quantitative evaluation methods seem believable, these measures just paint half the picture. One of the main contributions of proactive research in cybersecurity involve strengthening humans/individuals using the web and internet. Since humans are considered the weakest link in security research, we evaluate the efficacy of these fake emails by conducting a human evaluation study. 

We provide our 6 participants with a survey of 24 email bodies. For each email body, the participants are required to rate the quality of the email content based on three attributes - syntactical correctness, \textit{coherency} and \textit{fluency}. The scores for each of these attributes are based on a Likert scale $\in$ [1,5]. We also ask each participant to identify whether the email body is legitimate or not - which requires a yes/no response. Of the 24 email bodies, 12 emails are genuine or manually written and the rest are fake or system generated emails. Of the 12 system generated emails, we consider 7 emails which are generated by our sentence selection model (Generated$_{sentence}$) and 5 emails generated by our baseline word prediction model (Generated$_{word}$). We choose a balanced ratio to discern the effectiveness of our human participants in detecting fake emails from their genuine counterparts. 

Table~\ref{tab:humanevaluation} demonstrates the results of the human evaluation setup. We report the scores for syntax, coherency and fluency, averaged across all participants, on the combined set of system generated emails - Generated$_{all}$. We also report the same on the subset of emails generated by each of the baseline and deep generation models - Generator$_{word}$ and Generated$_{sentence}$ respectively, to compare the difference in system performance. We also report the scores on the combined set of system generated emails - Generated$_{all}$.
We further compare the statistical significance between the results on the manually written emails (Truth in Table~\ref{tab:humanevaluation}) with Generated$_{all}$ - syntax: 4.01 ($p-value=0.1377$), coherency: 3.31 ($p-value<10^{-5}$) and fluency: 3.19 ($p-value<10^{-5}$). We observe that while the difference in syntax scores between the legitimate and generated email bodies are not statistically significant,\footnote{$p-value$ calculated using the Unpaired Two Samples Wilcoxon test or Mann Whitney test}  the generated content is still behind in terms of coherency and fluency. However, in contrast, we observe that our human participants tend to have a low detection rate ($\approx$57\%) when it comes to generated emails, as observed in Table~\ref{tab:humanevaluation}, with the detection rate dropping to approximately 38\% in case of emails generated by the  sentence selection model. We also include the mean and standard deviation in the number of words ($\overline{W}$, $SD_{W}$) and sentences ($\overline{S}$, $SD_{S}$) in the generated and the manually written emails used in our survey. 

\begin{table*}[!htb]
\caption{Human evaluation results on the automated and true emails. Scores for Syntax (Syn), Coherency (Coh) and Fluency (Flu) vary between [1, 5]. Detection Rate has been reported as a percentage}
\label{tab:humanevaluation}
\centering
\begin{tabular}{|l|r|r|r|c|r|r|r|r|}
\hline
\multirow{2}{*}{\textbf{Email Content}} & \multicolumn{3}{c|}{\textbf{Scores}} & \multirow{2}{*}{\begin{tabular}[c]{@{}c@{}}\textbf{Detection} \\ \textbf{Rate}\end{tabular}} & \multirow{2}{*}{$\overline{\textbf{W}}$} & \multirow{2}{*}{\textbf{$SD_{W}$}} & \multirow{2}{*}{$\overline{\textbf{S}}$} & \multirow{2}{*}{$SD_{S}$}
\\
\cline{2-4}
                        & \textbf{Syn}     & \textbf{Coh}     & \textbf{Flu}     &  & &                                                                          &                    &                    \\ \hline
Generated$_{word}$              & 3.63    & 2.8     & 2.6     & 83.33                                                                      & 12     &       2.88     & 1.2      &  0.45        \\ \hline
Generated$_{sentence}$           & 4.29    & 3.67    & 3.62    & 38.09                                                                      & 13   &    6.21          & 2    &   0.81           \\ \hline
Generated$_{all}$          & 4.01    & 3.31    & 3.19    & 56.94                                                                      & 12   &   5           & 2      &    0.78        \\ \hline

Truth                   & 4.42    & 4.56    & 4.47    & 75                                                                         & 24    &      14.8       & 3      &   0.94         \\ \hline
\end{tabular}

\end{table*}

\section{Related Work}
The need to counter the active attacker issue has given rise to proactive research methods. While there exists classical techniques for phishing email detection~\cite{verma2012two},~\cite{chandrasekaran2006phishing},~\cite{basnet2008detection}, state-of-the-art malicious email detectors fail to detect a sophisticated or targeted attack. 
Riding on the wave of proactive research, many researchers have delved deeper into the realm of attack generation for different attack vectors but none as serious as emails. While the use of fully automated methods for text generation has been considered, there have not been much investigation into automatic modeling of coherent emails which can be used for targeted attacks.

\subsection{Natural Language Generation}
Use of deep neural networks have enabled building fully (or partially, with feature engineering,) automated models for natural language generation. From the perspective of written text, a substantially trained deep network is capable of emulating the writing style of an individual. This property has been leveraged in natural language research by making deep learners write a wide variety of text Shakespearean Sonnets~\cite{xiedeep}, poetry~\cite{ghazvininejad2016generating},~\cite{yi2018chinese}, answer generation~\cite{liu2018curriculum}.

While the use of grammar~\cite{baki2017scaling}, templates~\cite{chen2014two},~\cite{chen2014two2} and statistical language based models (e.g. N-grams~\cite{goodfellow2016deep}) are popular, Recurrent Neural Networks (RNNs) have been shown to be a more suitable choice owing to their ability to learn dependencies across the textual context~\cite{graves2013generating}. Long-Short Term Memory (LSTM) networks are more suitable for longer text sequences. However, while a fully automated system seems lucrative - controlling the coherency, topic and structure of the generated text can be quite challenging. 

Using RNNs for sequence-to-sequence learning is a popular practice for text generation as shown in~\cite{feng2018topic},~\cite{zhang2018reinforcing},~\cite{qian2018assigning},~\cite{jagfeld2018sequence}. Since, simple encoder-decoder architectures fail to model important or meaningful words and phrases, ~\cite{kiddon2016globally},~\cite{feng2018topic} use attention based encoder-decoder models for preserving coherence and context in the generated text. Other generation techniques include deep learning with Markov Models~\cite{wiseman2018learning}, variational auto-encoders~\cite{roberts2018hierarchical}, generative adversarial networks~\cite{press2017language}. While the aforementioned research works experiment with the architectures, such strategies are prone to generating incoherent content as the length of the generated text increases. The architecture used in this paper attempts at fixing this issue at a more global level - by implementing coherence in each sentence and then selecting the best sentence to be included in the generated text. 

\subsection{Attack Generation}
Growth in the field of proactive research has become more prominent in an effort to counter active attackers. Phishing is a largely unsolved cyber threat, worsened more by the proliferation in spear-phishing attacks. The ability of the perpetrators to deceive an individual by behaving as a legitimate entity can be automated for widespread social engineering attacks as studied in~\cite{nohlberg2008cycle} and~\cite{huber2009towards}. Researchers in~\cite{baki2017scaling},~\cite{DAS18.4},~\cite{giaretta2017community}, look at `weaponinizing' advanced machine learning techniques to launch sophisticated yet automated targeted attacks. While~\cite{baki2017scaling} uses a grammar-based approach for synthetic email generation; Das et. al.~\cite{DAS18.4} uses a more automated deep neural network for email generation, which suffers from incongruity in the generated context as shown by their evaluation. Other studies in an adversarial setting, which leverage natural language techniques, have been pursued in spreading malicious Twitter messages~\cite{seymour2016weaponizing}, generating malicious URLs~\cite{bahnsen2018deepphish}, generation of fake reviews~\cite{yao2017automated} as well as text messages~\cite{shropshire2018natural}. Automated means of synthesizing sophisticated attack vectors reduces the manual labor and provide phishers an opportunity to launch targeted attacks on a  much larger scale and magnitude. This in turn increases the chances of succeeding in an attack.

\subsection{Emails as Attack Vectors}
Emails are the most common and preferred method for social engineering attacks. \cite{drake2004anatomy} describes the \textit{modus operandi} and the structure of a common phishing email. Researchers have also delved deeper into the attributes and underlying psychological features that cause phishing and social engineering attacks to be successful in \cite{ferreira2017phish}, \cite{sheng2010falls}, \cite{giaretta2017community}. Techniques for automatic generation of synthetic emails have been discussed in \cite{baki2017scaling}, \cite{chen2014two}, but introducing attributes of deception into legitimate emails is a non-trivial task~\cite{DAS18.4},~\cite{carvalho2008modeling},~\cite{nohlberg2008cycle}.    

\section{Discussion and Conclusions}
We revisit \textit{two major error trends} observed in the evaluation of our word and character based generation models. \textit{First,} repetitions of tags and words in the generated text body.
A sample sentence generated by the word-based language model - ``The corres $ent$  $ent$  $ent$ Also $ent$ , $ent$ I we can operating a gift to ensure, are that extent will is a links are not $ent$" - demonstrates such behavior. While, we hypothesized such a behavior at larger text lengths, the brittleness in a model which uses characters or words as units for text generation can be observed for shorter text sequence generation as well. We believe, that the nature of the input on which the system is being modeled and the \textit{temperature} ($\tau$) parameter used for sample generation play an important role in this behavior of the predictive model.

While the RNN model generated text with `some' malicious intent in them - the examples shown above are just a few steps from being coherent and congruous. We designed an RNN based text generation system for generating targeted attack emails which is a challenging task in itself and a novel approach to the best of our knowledge. The examples generated however suffer from random strings and grammatical errors. We identify a few areas of improvement for the deep generation system - reduction of repetitive content as well as inclusion of more legitimate and phishing examples for analysis and model training. We would also like to experiment with addition of topics and tags like `bank account', `paypal', `password renewal', etc. which may help generate more specific emails. It would be interesting to see how a generative RNN handles topic based email generation problem.

\section*{Acknowledgements}
This research was supported in part by NSF grants CNS 1319212, DGE 1433817, DUE 1241772, and DUE
1356705. The study is also based upon work supported in part by the U. S. Army Research
Laboratory and the U. S. Army Research Office under contract/grant number W911NF-16-1-0422.

















\small{
\bibliographystyle{cys}
\bibliography{example}
}
\normalsize

\begin{biography}[]{} 
\end{biography}


\end{document}